\documentclass[10pt,twocolumn,letterpaper]{article}

\usepackage[pagenumbers]{cvpr} 

\usepackage[dvipsnames]{xcolor}

\usepackage{amsmath}
\usepackage{soul}
\usepackage[ruled,linesnumbered]{algorithm2e}

\definecolor{blue}{rgb}{0.21,0.49,0.74}
\usepackage[pagebackref,breaklinks,colorlinks,citecolor=blue]{hyperref}

\newcommand{\sysname}{MonoHair}
\newcommand{\methodname}{PMVO}

\newcommand\D{\Psi}
\newcommand\RawPointCloud{\mathcal{P}_{\text{raw}}}
\newcommand\LineMap{\mathcal{P}_{\text{out}}}
\newcommand\inner{ \mathcal{P}_{\text{in}}}
\newcommand\LineP{L^{\boldsymbol{p}}}
\newcommand\linetwod{l^{\boldsymbol{p}}}
\newcommand\point{\boldsymbol{p}}
\newcommand\uv{\boldsymbol{x}}
\newcommand\net{\mathcal{N}}
\newcommand\render{\mathcal{R}}
\newcommand\OriMap{\mathcal{O}}
\newcommand\ConfMap{\mathcal{C}}
\newcommand\DepthMap{\mathcal{D}}

\newcommand\blfootnote[1]{%
  \begingroup
  \renewcommand\thefootnote{}\footnote{#1}%
  \addtocounter{footnote}{-1}%
  \endgroup
}

\def\paperID{*****} %
\def\confName{****}
\def\confYear{****}

\title{MonoHair: High-Fidelity Hair Modeling from a Monocular Video}

\author{Keyu Wu$^{1}$ 
\quad Lingchen Yang$^{2}$
\quad Zhiyi Kuang$^{1}$
\quad Yao Feng$^{2,4}$
\quad Xutao Han$^{1}$\\
\quad Yuefan Shen$^{1}$
\quad Hongbo Fu$^{3,5}$
\quad Kun Zhou$^{1}$
\quad Youyi Zheng$^{1\dagger}$\\[1.5mm]
$^1$ Zhejiang University \quad
$^2$ ETH Zurich \quad
$^3$ City University of Hong Kong \\
$^4$ Max Planck Institute for Intelligent Systems \\
$^5$ Hong Kong University of Science and Technology
}

\begin{document}

\twocolumn[{%
\renewcommand\twocolumn[1][]{#1}%
\maketitle
\begin{center}
    \centering
    \captionsetup{type=figure}
    \includegraphics[width=0.95\linewidth]{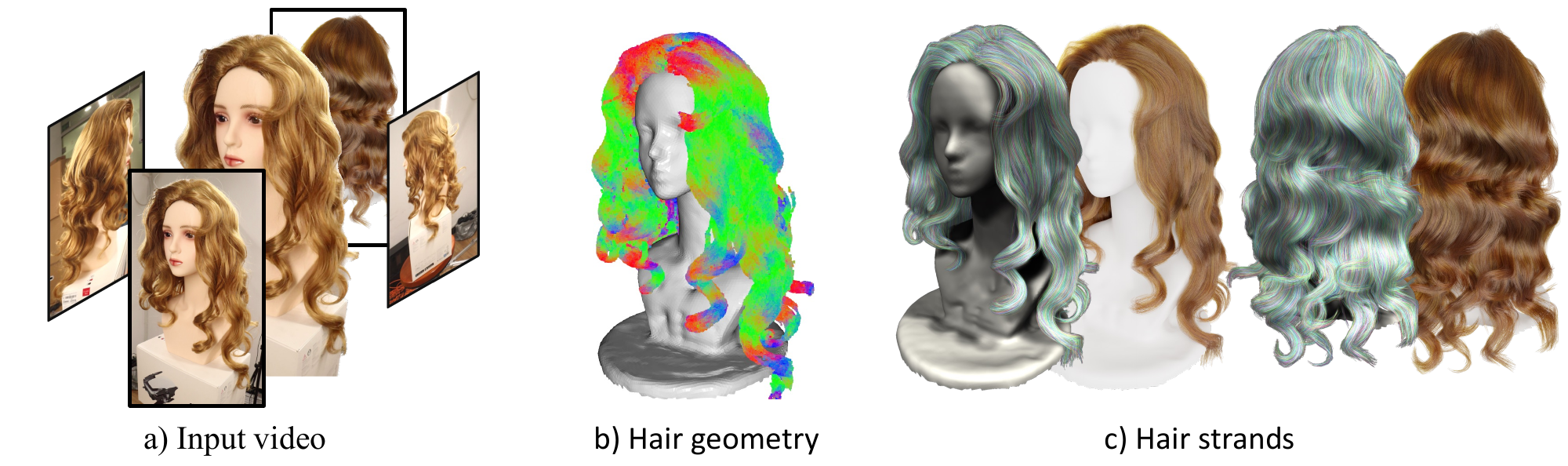}
    \vspace{-3mm}
    \captionof{figure}{
    We propose a generic framework for 3D hair modeling from monocular videos (a). It commences with a coarse raw geometry produced by volumetric representations. Subsequently, we extract the {exterior} geometry
    of the hair from the raw geometry and combine it with an inferred interior structure to obtain the complete 3D hair geometry (b). Finally, we %
    recover the corresponding 3D hair model at the strand level. Our method can reconstruct diverse hairstyles and achieve high-fidelity hair modeling results (c). 
    } \label{fig:teaser}
    \vspace{-1mm}
\end{center}%
}]

\maketitle
\blfootnote{$^\dagger$Corresponding author: Youyi Zheng.}

\begin{abstract}

 Undoubtedly, high-fidelity 3D hair is crucial for achieving realism, artistic expression, and immersion in computer graphics. While existing 3D hair modeling methods have achieved impressive performance, the challenge of achieving high-quality hair reconstruction persists: they either require strict capture conditions, making practical applications difficult, or heavily rely on learned prior data, obscuring fine-grained details in images. To address these challenges, we propose \sysname,a generic framework to achieve high-fidelity hair reconstruction from a monocular video, without specific requirements for environments. Our approach bifurcates the hair modeling process into two main stages: precise exterior reconstruction and interior structure inference. The exterior is meticulously crafted using our Patch-based Multi-View Optimization (\methodname). This method strategically collects and integrates hair information from multiple views, independent of prior data, to produce a high-fidelity exterior 3D line map. This map not only captures intricate details but also facilitates the inference of the hair’s inner structure. For the interior, we employ a data-driven, multi-view 3D hair reconstruction method. This method utilizes 2D structural renderings derived from the reconstructed exterior, mirroring the synthetic 2D inputs used during training. 
 This alignment effectively bridges the domain gap between our training data and real-world data, thereby enhancing the accuracy and reliability of our interior structure inference. Lastly, we generate a strand model and resolve the directional ambiguity by our hair growth algorithm. Our experiments demonstrate that our method exhibits robustness across diverse hairstyles and achieves state-of-the-art performance. For more results, please refer to our project page \url{https://keyuwu-cs.github.io/MonoHair/}

\end{abstract}   

\begin{figure*}[ht]
		\centering
		\includegraphics[width=0.95\textwidth]{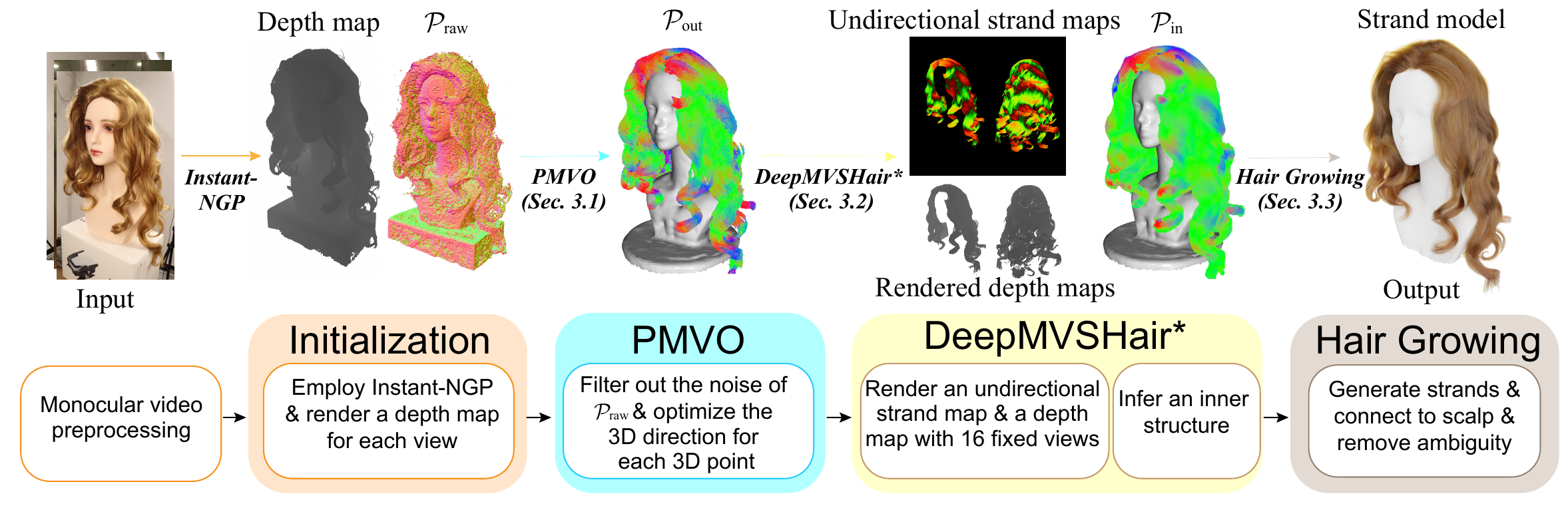}
        \vspace{-2mm}
		\caption{An %
  overview of our 3D hair reconstruction pipeline.}
		\label{fig:pipeline}
		\vspace{-3mm}
        \setlength{\intextsep}{-2mm}
\end{figure*}

\section{Introduction}
\label{sec:intro}

{Hair is a key feature in digital humans, and a detailed 3D hair model will undoubtedly enhance their realism \cite{cao2014face,hu2017avatar,hadap2007strands,li2015facial,luo2013structure}. However, hair modeling is an intricate endeavor, fraught with challenges at every turn. The high complexity stems from the unique geometry of individual strands twisting and turning in myriad ways.}

In computer graphics, 3D hair is commonly modeled as 3D strands, facilitating rendering and simulation processes. 
{Although accurate strand-based hair modeling can be realized using a light stage \cite{nam2019strand}, it relies on dense synchronized cameras. Besides, the reconstructed hair is incomplete since only the hair's exterior is captured %
neglecting its inner structure.}
To address these issues, several multi-view techniques \cite{zhang2017data,kuang2022deepmvshair} infer the inner structure based on an existing dataset \cite{hu2015single} while streamlining the experimental setup, for instance, by using sparse calibrated cameras. However, this sparsity in turn %
compromises the hair reconstruction quality.
Recently, 
\cite{sklyarova2023neural_haircut} utilizes dense views extracted from a casually-filmed video, achieving better quality and user-friendliness. However, they rely heavily on the data prior and tend to over-smooth the reconstruction results for certain specific hairstyles. This issue is particularly noticeable in hairstyles not well represented in the dataset, such as curly hair types.

{Similar issues exist with %
{the data-prior-based} multi-view methods. First, data priors, built by learning 3D hair generators with synthetic data, face diminished effectiveness when applied to real data, mainly due to the significant domain gap between the synthetic training data and real-world testing data. %
Second, the significant reliance on data priors tends to overshadow the rich information contained in the original {images}.
This neglect results in the prior over-dominance, preventing these methods from modeling %
fine-grained curly hair geometry, typically absent in the current dataset\cite{sklyarova2023neural_haircut,kuang2022deepmvshair,zhang2017data,zhang2019hair}. While such priors are essential for inferring plausible interior hair structures not readily discernible from input data, we believe that the hairstyle's outer layer can be modeled directly and used %
to enhance and refine the deduced interior structure.
}

{As shown in \cref{fig:teaser},} to address these concerns, we propose a generic solution for reconstructing hair from a monocular video. Our approach begins by initializing a coarse geometry through learning a Neural Radiance Field (NeRF), followed by sampling around the coarse geometry %
to generate a dense raw point cloud capable of representing diverse and complex hairstyles. To %
refine this raw point cloud, we introduce \emph{\methodname}, which leverages the rich information contained in the input video frames %
to reconstruct a high-quality hair exterior. This approach significantly alleviates the issue of prior over-dominance by focusing on the hair's exterior details. Lastly, to infer %
the missing inner hair structure, we adopt a data-driven, multi-view hair method, which %
takes 2D hair information as input. However, instead of applying Gabor filter on the images to get the input, we directly utilize 2D structural renderings derived from the reconstructed exterior, which mirrors the synthetic 2D inputs used during training. This alignment effectively bridges the domain gap between our training data and real-world testing data,
thereby enhancing the accuracy and reliability of our interior structure inference.

In summary, the main contributions of our work include:
	\begin{itemize}
		\item We propose a lightweight and generic framework for 3D hair {reconstruction} from a monocular video. This framework can robustly reconstruct diverse hairstyles, including curly hair, and outperforms state-of-the-art monocular video based hair reconstruction method in reconstruction quality. %
  It also offers a speed improvement of more than tenfold.
		\item We present a novel process for extracting high-quality exterior hair structures from noisy coarse geometries. This is achieved through our innovative patch-based multiview optimization, incorporating two novel cost functions: ray regularization and patch-wise angular loss.
		\item We introduce an undirectional strand map, generated by rendering the high-quality hair exterior, to bridge the gap between synthetic and real-world data. This enhances the reliability of interior hair structure inference.
  
	\end{itemize}

\section{Related work}
\label{sec:Related_work}

\paragraph{Implicit Representations in 3D Hair Modeling.} 
Recently, implicit representations such as NeRF %
\cite{mildenhall2020nerf,muller2022instant} and implicit surfaces~\cite{yariv2020multiview,wang2021neus} have emerged and gained extensive use in novel view synthesis \cite{barron2021mip,barron2022mip,weng2022humannerf,mildenhall2020nerf} and general scene reconstruction \cite{oechsle2021unisurf,wang2023neus2,fu2022geo}. Their primary advantages are high-quality results %
without meticulous camera calibration, and their flexibility for modeling various structures, including hair. For instance, studies like ~\cite{wang2022hvh,wang2021learning,wang2023neuwigs,rosu2022neural,Feng2023DELTA} employ volumetric representations to implicitly capture hair from multi-view images or monocular videos. However, the hair models generated by these methods fall short of meeting the standards for high-quality 3D strand-based hair reconstruction. Despite this limitation, such representations are valuable for providing initial coarse geometry.

\paragraph{Optimization Based Hair Reconstruction.} 
In the early stages of hair reconstruction research, strand-based representations were the primary focus, beginning with the pioneering work by Paris et al. \cite{paris2004capture}. This approach set the stage for numerous optimization-based hair reconstruction methods. For instance, Luo et al. \cite{luo2012multi, luo2013structure} optimized a hair mesh to incorporate fine-grained details using hair orientations as constraints. Similarly, other studies \cite{luo2013wide, hu2014robust} employed Multi-View Stereo (MVS) techniques to generate point clouds, optimizing shape primitives like strands and ribbons to create complete hair models.
Building upon these methods, Nam et al. \cite{nam2019strand} introduced a line-based PatchMatch MVS approach, capable of reconstructing high-precision hair segments using a dense capture setup with synchronized cameras. However, their method requires extensive multi-view calibrations and specific lighting conditions, posing practical challenges for average users. Moreover, the reconstructed hair models lack interior structures, thus significantly limiting %
their applicability.

\paragraph{Hair Reconstruction with Data Prior.}
Since the USC-HairSalon dataset \cite{hu2015single} was released, using data priors for 3D hair reconstruction has gained widespread popularity. Studies like \cite{chai2016autohair,zhang2017data} have developed data-driven methods that select and modify hairstyles from a database to match them with the geometry seen in images.
Further, research illustrated in \cite{wu2022neuralhdhair,zheng2023hairstep,yang2019dynamic,zhou2018hairnet,saito20183d,zhang2019hair} has shown how single images can be input into neural networks trained with synthetic data to create strand-based hair models. These deep learning methods also work well with a few images from different views, as shown in \cite{kuang2022deepmvshair}. However, they struggle to accurately reconstruct hairstyles not present in the database. Additionally, their effectiveness is reduced when applied to real data, primarily due to the significant domain gap between synthetic training data and real-world testing data. %
Recently, \cite{sklyarova2023neural_haircut} introduced a novel method that improves hair strand reconstruction by integrating various techniques. This method excels in mitigating issues associated with using synthetic data and produces impressive results. However, it heavily relies on data priors and is less effective in reconstructing curly hair. {Please refer to \cref{compare} for the discussion.}

\section{Method}
\label{Method}

\cref{fig:pipeline} shows the pipeline of \sysname.
Starting with a set of images $\{I\}$, uniformly sampled from a captured monocular %
video, we employ NeRF %
\cite{muller2022instant,mildenhall2021nerf,barron2021mip} 
for scene reconstruction. This yields a raw point cloud, $\RawPointCloud$, which represents the hair region, albeit in a noisy manner. Given $\RawPointCloud$, we aim to reconstruct an accurate exterior layer {$\LineMap$} and infer a plausible interior structure {$\inner$}. Based on these two components, we then generate hair strands.

Our \methodname~(denoted as $\D$) steps in to carve out a clear hair exterior structure. By leveraging the information from images $\{I\}$, it refines $\RawPointCloud$ and associates each point around the hair's boundary with a 3D hair-growing direction. This process results in a 3D line map, $\LineMap = \D( \RawPointCloud,I)$, where each 3D line can be {represented} by its
3D position $\point$ and 3D direction $\boldsymbol{d}$: $\LineP = \left\{\point,\boldsymbol{d}\right\}$ (\cref{Surface Optimization}). Notably, $\LineMap$ provides a structured representation of the hair's outer region, enabling the generation of coherent and high-quality 2D hair structure renderings. 

Subsequently, we employ a hair generation network DeepMVSHair \cite{kuang2022deepmvshair} {and improved it to adjust to our pipeline}, {denoted as DeepMVSHair* ($\net$). This network }is pre-trained on a synthetic 3D hair database and takes in the {2D undirectional strand maps} derived from $\LineMap$ to deduce the hair's intricate inner structure, represented as { $\inner = \net(\render( \LineMap))$ (\cref{infer_inner}), {where $\render$ denotes the rendering operation}. }

Finally, our strand generation module $\zeta$ extracts the %
hair strands $S$ from the reconstructed outer layer $\LineMap$ and the inferred inner structure $\inner$ (\cref{Strands_generate}). This can be formulated as: %
\begin{equation}
    S = \zeta(\LineMap,\inner).
\end{equation}

\subsection{Patch-based Multi-View Optimization}\label{Surface Optimization}

As discussed in {\cref{sec:intro}}, a fine-grained exterior structure is essential for hair reconstruction. However, obtaining such a %
high-precision exterior hair structure %
is nontrivial. Inspired by \cite{nam2019strand}, integrating multi-view image information to produce a 3D line map $\LineMap$ is a possible solution. 
However, their method reconstructs the exterior from scratch with dense calibrated images and %
is highly limited by expensive capture equipment and strict capture conditions. 
In response to these limitations, Neural Haircut \cite{sklyarova2023neural_haircut} employs NeuS \cite{wang2021neus} to first initialize a coarse 3D hair geometry and then refines it %
through differentiable rendering and a learned data prior. However, this approach applies the data prior to the overall hair shape, causing a loss of fine-grained hair details. Additionally, the representation of hair geometry using a signed distance field (SDF) further exacerbates the over-smoothness.
Thus, our method reconstructs the exterior geometry of hair without any hair data prior constraint, preserving as {many details} from the image as possible. 

\begin{figure}[t]
		\centering
		\includegraphics[width=\linewidth]{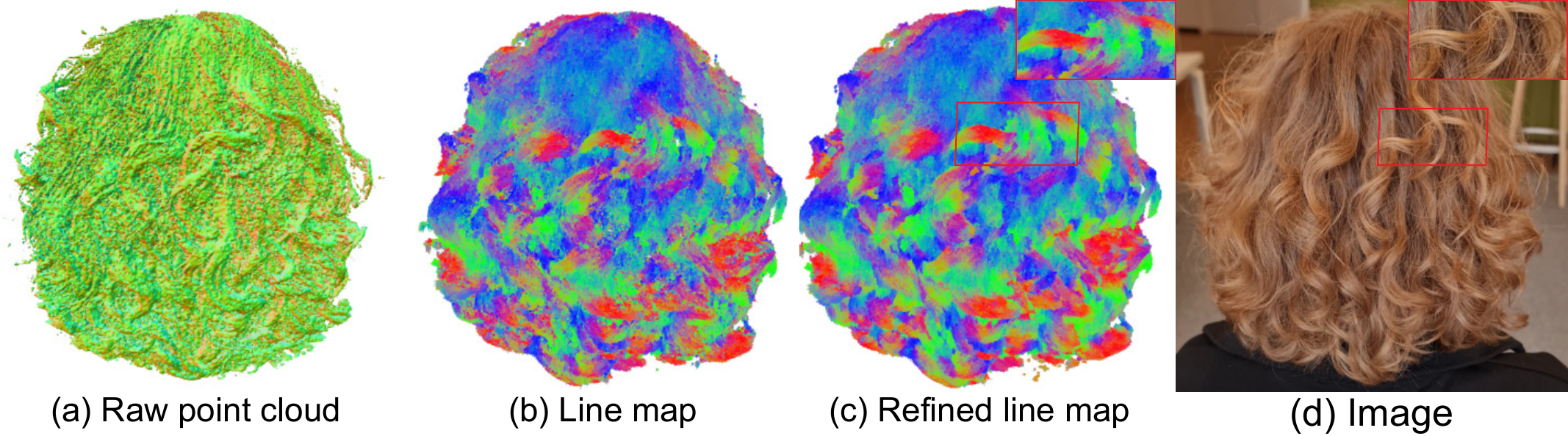}
		\caption{Visualization of the line map extracted from raw geometry.}
		\label{fig:data}
		\vspace{-4mm}
\end{figure}

Specifically, we first initialize the coarse hair structure using the point cloud extracted from NeRF \cite{barron2021mip,mildenhall2021nerf,muller2022instant} instead of NeuS \cite{wang2021neus}.
For efficiency, we utilize the Instant-NGP framework \cite{muller2022instant} and obtain an initial coarse geometry by applying a threshold of 2.5 to the learned density values. Subsequently, we densely sample around the obtained coarse geometry %
to obtain $\RawPointCloud$. Our key observation is that although $\RawPointCloud$ is very noisy and even terrible, it encompasses nearly all of the hair's exterior geometry, as illustrated in \cref{fig:data} (a). Thus, our proposed \methodname~  is designed to first eliminate the noise (points not belonging to the hair) while preserving the fine-grained geometry of the hair and then calculate a 3D growing direction at each remaining point.

\textbf{Input Data.} The input of our \methodname~consists of the coarse point cloud $\RawPointCloud$ and an image set, uniformly sampled from the captured monocular video. Subsequently, for each image, we obtain three maps: a %
2D orientation map $\OriMap$, a %
confidence map $\ConfMap$, and a %
depth map $\DepthMap$. {The 2D orientation map $\OriMap$ and the confidence map $\ConfMap$ are extracted using the Gabor filter similar to the previous work \cite{paris2004capture}. $\OriMap$ contains the 2D hair growth direction at each pixel %
while $\ConfMap$ measures the confidence of the estimated direction. The depth map $\DepthMap$ is obtained by directly rendering the depth of $\RawPointCloud$ with the estimated camera parameters from COLMAP \cite{schonberger2016structure}. $\DepthMap$ serves as the cue for judging the visibility $V^{\point}$ of point $\point\in{\RawPointCloud}$ by: $V^{\point}=1-\frac{\point_z - \DepthMap(\Pi(\point))}{\tau}$, where $\point_z$ is the $z$ coordinate of $\point$ and $\Pi$ is the projection function. $\tau$ represents a visible threshold, which we set to 5mm in our experiment to differentiate between the exterior (visible regions) and inner regions (invisible regions) of the hair. These maps will be used to calculate our cost function, as described below.}

\textbf{Cost Function.} For each point $\point\in{\RawPointCloud}$, our \methodname~attempts to find a %
correct %
3D line $\LineP$ by integrating information from all views in which $\point$ is visible. {Here, the correct 3D line should minimize our proposed cost function $\mathcal{L}_{opt}$, which}
describes the similarity between the projected 2D line $\linetwod$ of $\LineP$ and the 2D orientation in %
the corresponding views as follows:
\begin{equation}
    \mathcal{L}_{opt}(\point,\LineP) = \frac{\sum_{i=1}^N w_ig_i(\OriMap_{i}(\Pi_i(\point)),\linetwod_{i})}{ \sum_{i=1}^N w_i}, 
\end{equation}
where $i$ denotes a specific view, $\linetwod_i=\Pi_i(\LineP)$, $N$ is the number of views used, and {$w_i = V^{\point}_i \cdot \ConfMap_i(\Pi_i(\point))$}
is the weight of $\point$ in the $i$th view. $g_i(\OriMap_i(\Pi(\point)), \linetwod_i)$ 
is our {proposed {patch-wise} angular loss function}. This function measures the angle %
difference between the 2D orientation $\OriMap_i(\Pi_i(\point))$ corresponding to point $\point$ and the projected 2D line direction $\linetwod_i$ in the $i$th frame.
The key idea for extracting the hair line map $\LineMap$ from the raw point cloud $\RawPointCloud$ is that each 3D line $\LineP$ belonging to the hair can be projected into all visible views, and its 3D direction projections align with the corresponding 2D orientations %
in those views—a characteristic that noise does not possess. Thus, our objective is to find the best-fitting $\LineP$ that minimizes the cost function $\mathcal{L}_{opt}$:
\begin{equation}
    \LineP = \underset{\LineP}{\arg\min}~\mathcal{L}_{opt}(\point,\LineP).
\end{equation}

\begin{figure}[t]
		\centering
		\includegraphics[width=\linewidth]{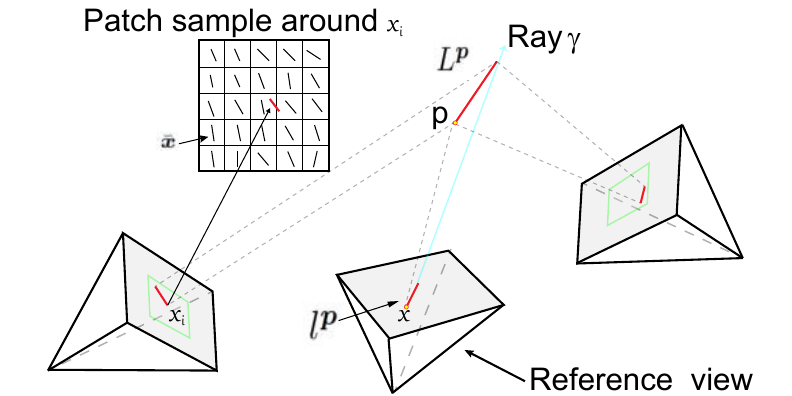}
		\caption{{Schematic diagram of patch-based multi-view optimization.} }
		\label{fig:sample}
		\vspace{-6mm}
\end{figure}

\textbf{{Patch-wise} Angular Loss.} 
Our patch-wise angular loss aims to measure the re-projection cosine difference between the projected 2D line and the corresponding 2D orientation.
{As shown in \cref{fig:sample},} for the 2D projection $\uv = \Pi(p)$ of $\point$ at each frame, we sample $k$ number of 2D points centered on $\uv$, $k=r^2$, where r is a patch size (set as 5 pixels in our experiments). Then, we can formulate our patch-wise angular loss $g_i$ as:
\begin{equation}
    g_i(\OriMap_{i}(\uv),l^{\point}_{i}) = \sum_{\Bar{\uv} \in X_i(\uv)}
    \ConfMap_{i}(\bar{\uv})\cdot(1- cos(\OriMap_i({\Bar{\uv}}),l^{\point}_{i})),
\end{equation}
where $X_i$ is a 2D point set sampled 2D centered on $\uv$ in the $i$th view. This loss function has %
two advantages. %
First, it allows some calibration errors. Second, it can also smooth the direction of locally adjacent 3D lines, the same as the properties of hair.

\textbf{Optimization.}
{As shown in \cref{fig:sample}},  to solve the cost function $\mathcal{L}_{opt}$ robustly, we first select a reference frame with the largest confidence in the frames where $\point$ is visible to initialize $\LineP$ with a 3D vector $\LineP = (\OriMap(\Pi(\point)),0)$.
It's easy to find that the correct 3D line $\LineP$ should intersect with the ray $\gamma$ emitted from the other end of the 2D line $\linetwod$ at the reference view. Then, in our process of optimizing $\LineP$, we constrain the direction of $\LineP$ by the following regularization:
\begin{equation}
    \mathcal{L}_{reg} = dist(\gamma, Ray(\point,\LineP)), 
\end{equation}
where $Ray(\point,\LineP)$ represents the ray starting from point $\point$ along the $\LineP$ direction, and $dist(\cdot)$ is the function for calculating the closest distance between two rays.

\begin{figure}[t]
		\centering
		\includegraphics[width=\linewidth]{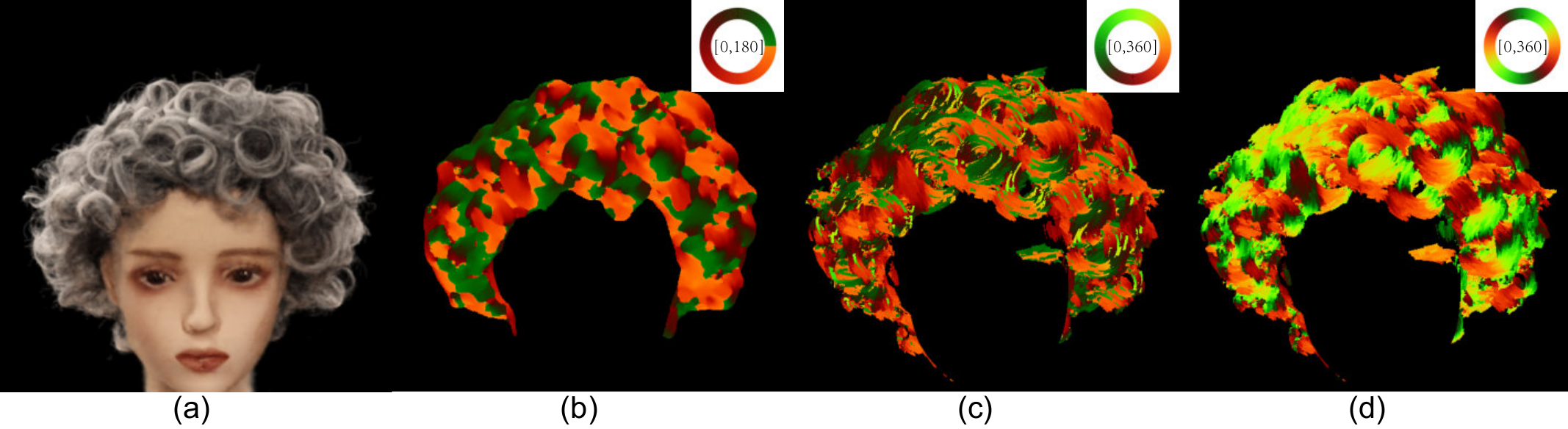}
        \vspace{-6mm}
		\caption{Visualization of different 2D hair growth direction maps. (a) Portrait image. (b) Orientation map reported in \cite{kuang2022deepmvshair}. (c) Rendered strand map. (d) Rendered undirectional strand map.
  }
		\label{fig:ori}
		\vspace{-5mm}
\end{figure}

\begin{figure*}[!ht]
		\centering
		\includegraphics[width=0.95\textwidth]{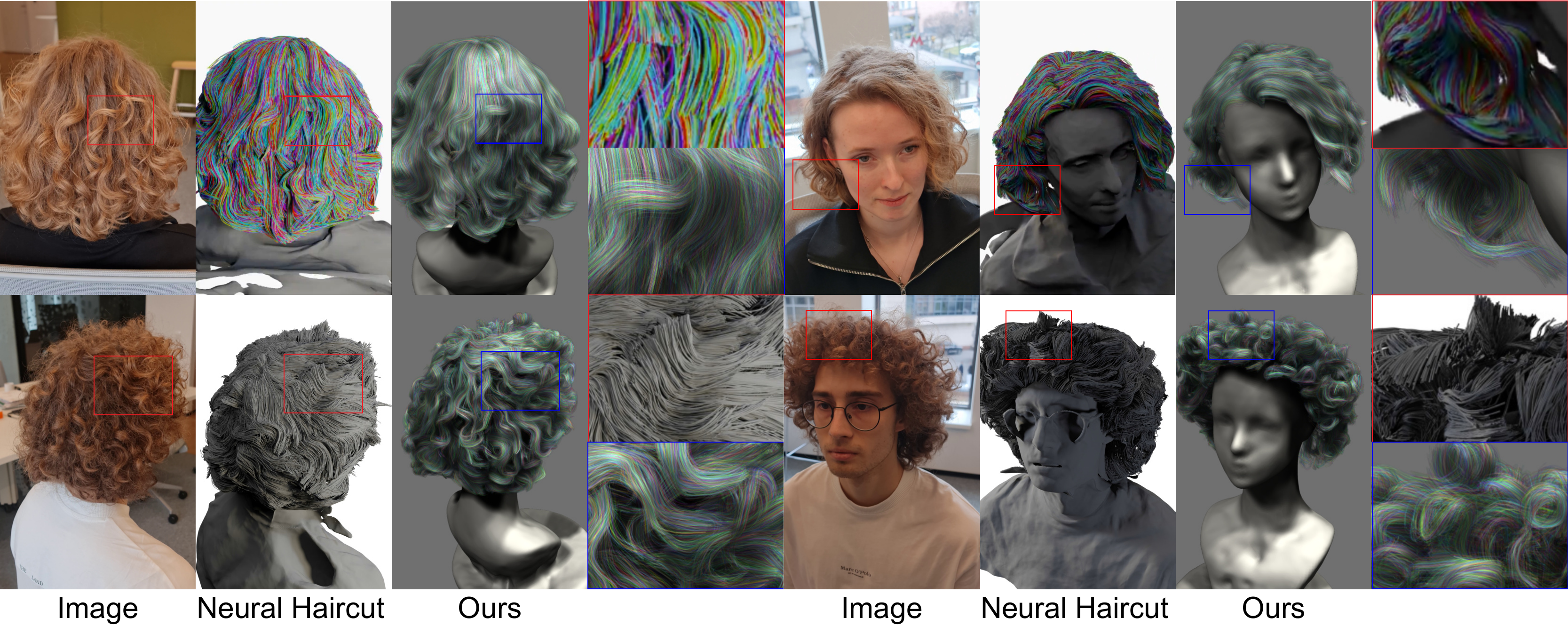}
		\caption{Qualitative comparison with Neural Haircut \cite{sklyarova2023neural_haircut}, which %
  has limited ability to reconstruct complex hairstyles, especially for curly hair (the reconstruction results are all from their paper). On the contrary, our reconstruction results can maintain more details in the images. }
		\label{fig:Compare}
		\vspace{-4mm}
\end{figure*}

\textbf{Line Refinement.}  For each $\point \in \RawPointCloud$, we filter %
out the inner points, which are not visible in all frames. Then we keep the point $\point$ with $\mathcal{L}_{opt}(\point,\LineP)<0.05$ (about 15 degrees) as the final $\LineMap$, as shown in \cref{fig:data} (b). However, the %
extracted hair's exterior geometry suffers from %
some noise. Therefore, we find 100 neighbors of $\point$, denoted as $\{\point_{nei}\}$, %
and calculate the variance between $\point$ and $\{\point_{nei}\}$: %
\begin{equation}
    var(\point) = cos(\LineP,\boldsymbol{avg}(\mathcal{L}^{\point_{nei}})).
\end{equation}
Subsequently, we update the $\LineP$ with $\boldsymbol{avg}(\mathcal{L}^{P_{nei}})$ if $var(\point)>0.015$ (about 10 degrees) to produce the final exterior of hair $\LineMap$, as shown in \cref{fig:data} (c).

\subsection{Infer Interior Geometry}\label{infer_inner}
For a complete acquisition of hair geometry, inferring the hair's inner structure is necessary. We employ a method that incorporates data priors, similar to DeepMVSHair\cite{kuang2022deepmvshair}. %
{Their method {takes multi-view calibrated images as input and} trains a HairMVSNet on a synthetic dataset to integrate multi-view hair structure features to infer the hair geometry, represented as a pair of a %
3D occupancy field and a %
3D orientation field.}
However, the direct application of this approach to our problem faces two challenges: 1) the need for calibrated images and 2) a domain gap between synthetic and real 2D data. The 2D orientation maps, extracted using Gabor filters from images, are limited by capture quality and thus introduce directional ambiguity, a common issue in 3D hair modeling.

To overcome these limitations, we propose two improvements
: 1) Instead of extracting 2D orientation maps from calibrated images, we render the extracted exterior layer of hair, \(\LineMap\), to 16 fixed synthetic views (\textbf{DeepMVSHair*} is trained using these fixed 16 views), as shown in \cref{fig:ori}. 
{Our key observation is that independently extracting its own 2D orientation map from each image is easily affected by image quality, viewing angle, and occlusion. In contrast, integrating geometric information from multi-view images into 3D and rendering it into 2D results in clearer geometry and greater robustness. Furthermore, this process is the same as the training data preparation.} Additionally, %
this strategy facilitates the rendering of additional views, enabling us to produce images from numerous angles and positions while ensuring the precision of camera parameter settings. 2) While \cite{zheng2023hairstep}  attempted to train a neural network to mitigate the ambiguity only in the frontal view, it is insufficiently robust for other views. Our observations suggest that resolving ambiguities in the 3D space is often simpler than directly in the image space. Therefore, we tackle this issue in a subsequent stage (\cref{Strands_generate}) by defining a 2D undirectional strand map $\mathcal{U}$ as:
\begin{equation}
    \mathcal{U} = (\cos(2\cdot{\mathcal{O}}), \sin(2\cdot{\mathcal{O}})),
\end{equation}
where \(\mathcal{O}\in[0,180]\). Here, \(\mathcal{O}\) and \(\mathcal{O}+180\) are encoded into the same color space to better facilitate the training of DeepMVSHair \cite{kuang2022deepmvshair}. Concurrently, we render a depth map \(\mathcal{D}\) for each view using our high-quality exterior hair structure. Consequently, we can infer the inner structure using the improved \textbf{DeepMVSHair*} as follows:
\begin{equation}
    \inner = \net(\LineMap, \mathcal{U}, \mathcal{D}).
\end{equation}

Besides, to accommodate the aforementioned modifications, we also design a new loss function 
for their orientation prediction component. Specifically, the predicted 3D direction $d$ and its opposite direction $-d$ are considered to be the same direction. This can be formulated by:
\begin{equation}
    \mathcal{L}_{ori} = \frac{1}{N}\sum_{i}^{N}\min(\frac{||\hat{d}-d||_1}{3},\frac{||\hat{d}+d||_1}{3}), 
\end{equation}
where $N$ is the number of views, $\hat{d}$ is the ground truth. For more details, please refer to \cite{kuang2022deepmvshair}.

\subsection{Strand Generation} %
\label{Strands_generate}
To generate a complete 3D strand model, we need to merge $\LineMap$ and $\inner$. Specifically, for each point $\point$ in $\inner$, we selectively integrate the points that are invisible in all views, combining them with %
$\LineMap$ to form our final hair geometry $\mathcal{H}$. This strategy ensures that the data prior does not overshadow the details in the hair's exterior geometry. Subsequently, we voxelize the space and convert %
$\mathcal{H}$ to a high-resolution 3D orientation field and then
use forward Euler and backward Euler to generate segments $\left\{s\right\}$ similar to previous works \cite{kuang2022deepmvshair,wu2022neuralhdhair}. The difference is that we recursively connect short segments into long strands instead of connecting to the scalp root directly {since it is difficult to distinguish which end of a segment is the root when the short segments are close to the scalp.} Besides, we connect long strands to the scalp and detect the direction of unconnected strands using connected strands to resolve %
direction ambiguity. \cref{grow} presents our growing step in detail.

\setlength{\intextsep}{10pt}
\begin{algorithm}
\caption{Strand Generation} %
\label{grow}   
\textbf{Input:} Hair scalp $\Omega$, hair segment set $\{s\}$ \\
\textbf{Output:} Strands connected to scalp $S^c$ \\
\textbf{Step1:} For each segment $s$, find the nearest neighbor segment to root ($s^r_{nei}$) and tip ($s^t_{nei}$) respectively.\\
\textbf{Step2:} Recursively connect ($s^r_{nei}$) and ($s^t_{nei}$) to produce long strand set $\{s^l\}$\\
\While{len($\{s^l\}$)$\neq$ 0}{
\textbf{Step3:} For each $s^l$, calculate the distance $dist(s^l,\Omega)$ of its end (either root or tip) closest to the scalp. If $dist(s^l,\Omega)<15 mm$, include $s^l$ to the set $S^c$ and remove it in $\{s^l\}$. \\
\textbf{step4:} For each $s^l$, find the neighbor $s^c$ in $S^c$ with a distance $dist(s^l,s^c)<5mm$, having the most similar growth direction, then resolve the ambiguity in growth direction based on it. Where $s^c \in{S^c}$.\\
\textbf{Step5:} For the remaining $s^l$, find the neighbor $s^c$ in $S^c$ with a distance $dist(s^l,s^c)<2mm$, and connect it with the closest point on $s^c$.}
\vspace{-1.7mm}
\end{algorithm}

\begin{figure}
    \centering
    \resizebox{\linewidth}{!}{
    \begin{tabular}{l rrr | rrr | rrr}
        \setlength{\tabcolsep}{0pt}
        & \multicolumn{9}{c}{\textbf{Thresholds: mm} $/$ \textbf{degrees}} \\
        \textbf{Method} & $2 / 20$ & $3 / 30$ & $4 / 40$ & $2 / 20$ & $3 / 30$ & $4 / 40$ & $2 / 20$ & $3 / 30$ & $4 / 40$ \\
        \cline{2-10}
        & \multicolumn{3}{c}{\textbf{Precision}} & \multicolumn{3}{c}{\textbf{Recall}} & \multicolumn{3}{c}{\textbf{F-score}} \\
        \hline
        DeepMVSHair~\cite{kuang2022deepmvshair}	&	43.9	&	67.2	&	79.5	&	9.2	&	19.5	&	24.8	&	15.2	&	30.2	&	37.8	\\
 DeepMVSHair*	&	49.5	&	77.1	&	86.3	&	9.3	&	19.0	&	25.1	&	15.7	&	30.5	&	38.9	\\
 Neural Haircut~\cite{sklyarova2023neural_haircut}	&	52.9	&	78.1	&	88.4	&	9.8	&	17.8	&	\textbf{26.3}	&	16.4	&	28.7	&	40.3	\\
 Ours	&	\textbf{60.8}	&	\textbf{83.3}	&	\textbf{92.1}	&	\textbf{10.4}	&	\textbf{19.3}	&	25.9	&	\textbf{17.8}	&	\textbf{31.3}	&	\textbf{40.4}	\\

    \end{tabular}
    }
    \vspace{-0.3cm}
    \captionof{table}{Quantitative comparison with \cite{kuang2022deepmvshair,sklyarova2023neural_haircut}. Our method achieves the %
    highest precision and F-score.}
    \label{tab:comparison}
     \vspace{-0.4cm}
\end{figure}

\begin{figure}
    \centering
        \resizebox{\linewidth}{!}{
        \begin{tabular}{c|c| c| c| c|}
        
            Method & coarse geometry & fine geometry & strand generate & total  \\
          \hline
            Neural Haircut \cite{sklyarova2023neural_haircut} & 24-36h & 48-72h & / & 72-108h \\  \hline
            Ours & ~5min & 3-4h & 1-2h & 4-6h\\
          \hline
        \end{tabular}
        }      
    \vspace{-0.2cm}
    \captionof{table}{Comparison with Neural Haircut 
    \cite{sklyarova2023neural_haircut} in terms of time consumption. Our method is ten times faster.}
    \label{tab:time_compare}
     \vspace{-0.4cm}
\end{figure}

\begin{figure*}[t]
		\centering
		\includegraphics[width=0.9\textwidth]{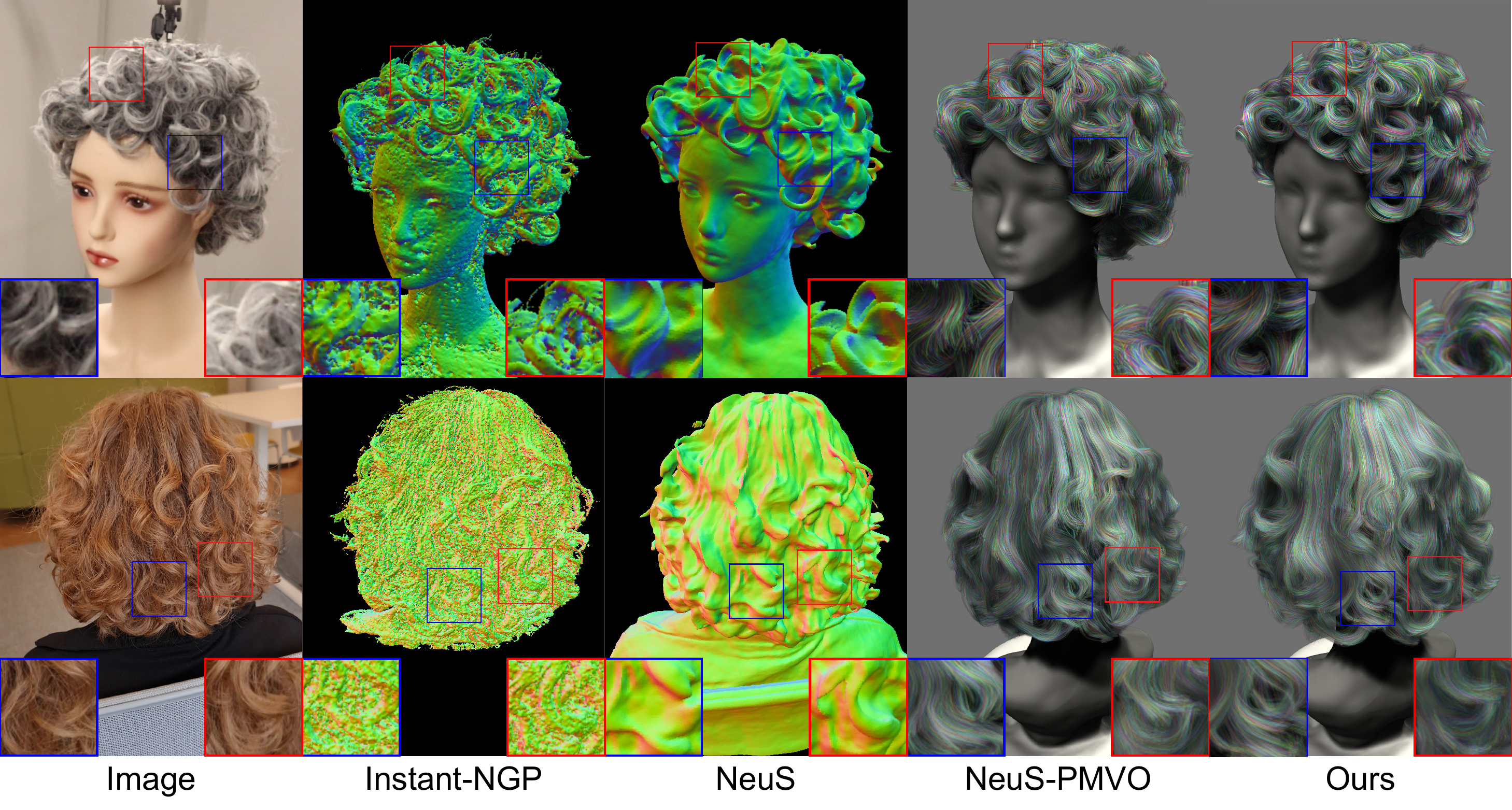}
  	\vspace{-2mm}
		\caption{Qualitative comparison with Instant-NGP \cite{muller2022instant} and NeuS \cite{wang2021neus}. These %
  volumetric approaches can only produce coarse hair geometry. We also compare the results with {another initialization method (NeuS-\methodname)}. Although they yield similar results, initializing coarse geometry using NeuS tends to obscure %
  fine-grained details. 
  }
		
		\label{fig:compare_neus}
			\vspace{-3mm}
        \setlength{\abovecaptionskip}{8mm}
\end{figure*}

\section{Evalution}
\label{Evalution}
We train our improved DeepMVSHair* model on USC-HairSalon \cite{hu2015single}, which includes 343 hairstyles aligned with a template head. To augment the dataset, we applied random translations, rotations, and scaling, resulting in a total of 2,744 strand models. We evaluate our method through a comprehensive evaluation, encompassing both quantitative and qualitative comparisons, using synthetic data \cite{yuksel2009hair} and public real-world H3DS dataset \cite{ramon2021h3d} as well as real-world monocular video captures (\cref{compare}). We also conduct an ablation study to evaluate the importance of each component of our method (\cref{ablation}). Implementation details and more experiments please refer to supplementary materials.

\begin{figure}[t]
		\centering
		\includegraphics[width=\linewidth]{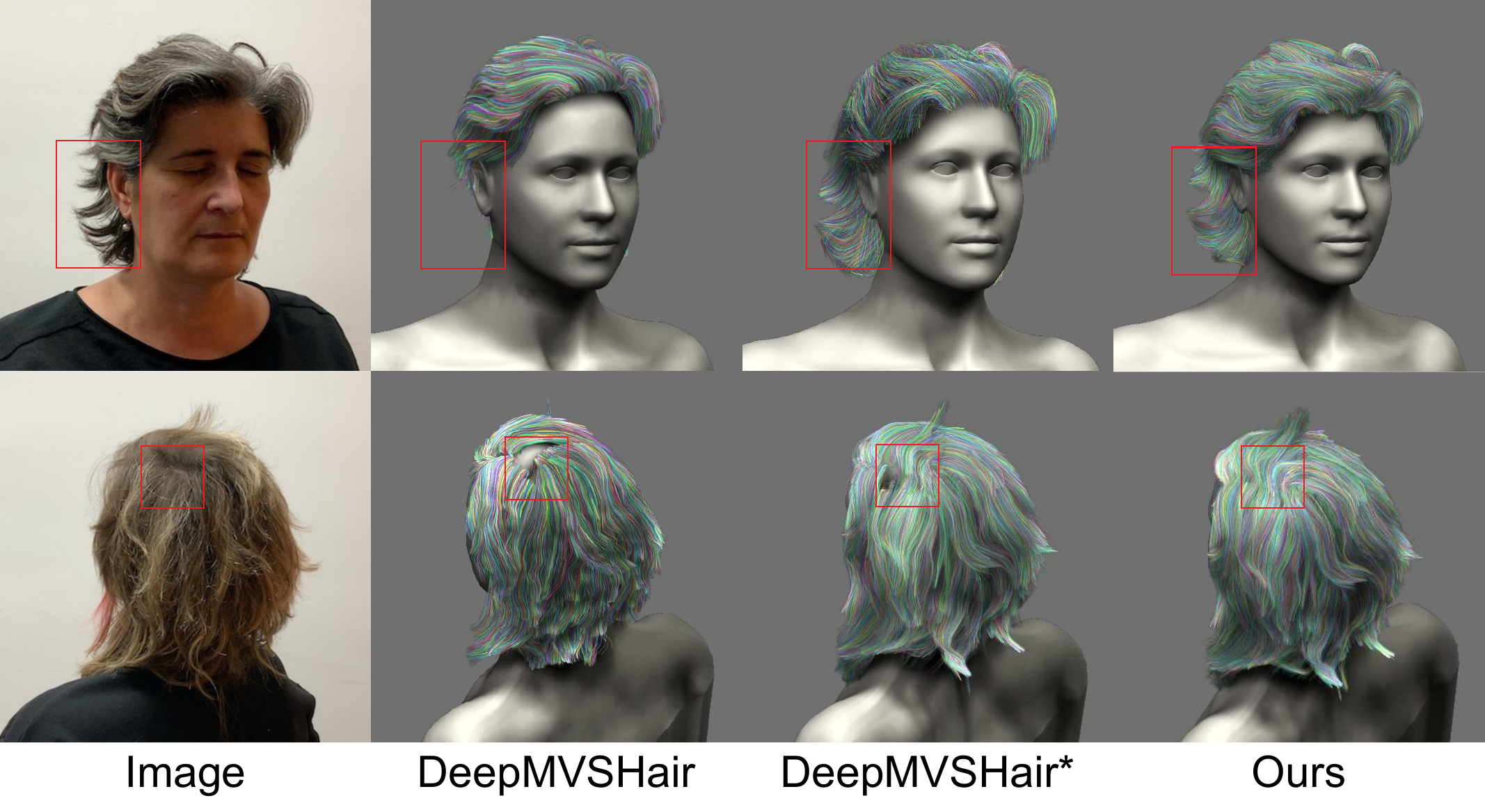}
        \vspace{-4mm}
		\caption{Qualitative comparison with DeepMVSHair \cite{kuang2022deepmvshair} on the %
  H3DS\cite{ramon2021h3d} dataset. Since the 2D orientation maps extracted from the images may be inconsistent across different views, their results %
  cause some lost geometry. %
  Our improved DeepMVSHair* effectively addresses this issue, though %
  there are still some details lost due to data prior limitations. {In contrast, our method escapes these limitations and achieves a high-fidelity result with richer details.} 
  }
		
		\label{fig:compare_mvs}
		\vspace{-6mm}
\end{figure}

\subsection{Comparison}\label{compare}
\textbf{Baselines.} We compare \sysname with strand-based hair modeling methods \cite{sklyarova2023neural_haircut,kuang2022deepmvshair,zheng2023hairstep,wu2022neuralhdhair}, as well as a popular 3D reconstruction method \cite{wang2021neus} and a NeRF-based method \cite{muller2022instant}. Where \textbf{Neural Haircut} \cite{sklyarova2023neural_haircut} reconstructs a %
strand model from a monocular video, consistent with our input. \textbf{DeepMVSHair} \cite{kuang2022deepmvshair} is a strand-based reconstruction method based on sparse multi-view images. In our evaluation, we compare our method with basic \textbf{DeepMVSHair} and our improved implementation (\textbf{DeepMVSHair*}), using both synthetic \cite{yuksel2009hair} and real-world data. \textbf{NeuS} \cite{wang2021neus} is a representative reconstruction method based on an SDF representation, while \textbf{Instant-NGP} \cite{muller2022instant} is commonly used for novel view synthesis. We compare our method with these methods and \textbf{NeuS-\methodname}, where \textbf{NeuS-\methodname} is our method with the SDF representation as coarse initialization geometry. Finally, we also provide comparisons with single-view based methods \textbf{NeuralHDHair}\cite{wu2022neuralhdhair} and \textbf{HairStep} \cite{zheng2023hairstep} in the supplementary materials.

\begin{figure*}[t]
		\centering
		\includegraphics[width=0.95\textwidth]{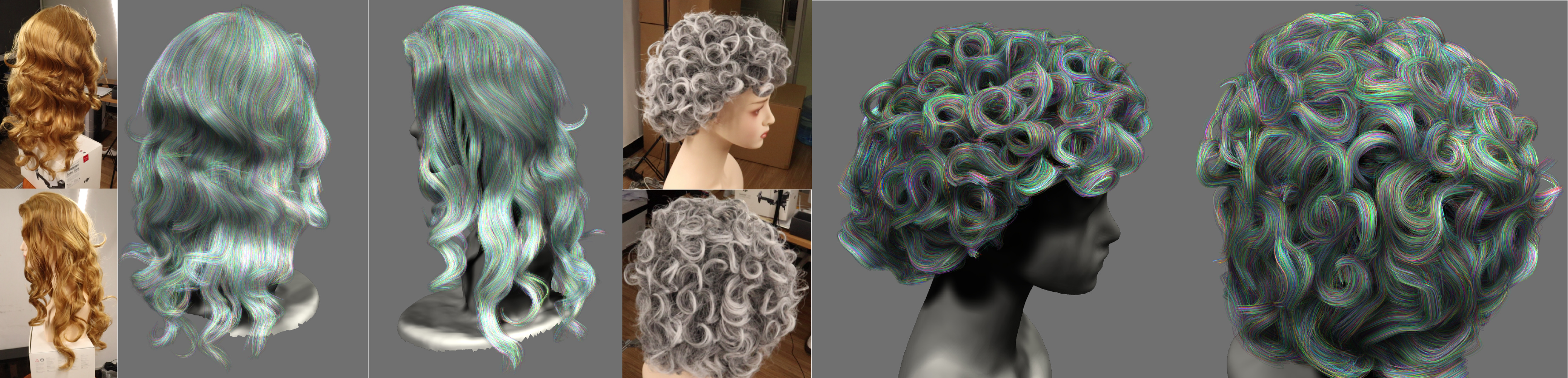}
		\caption{Given a monocular video, our method can reconstruct a high-fidelity strand model, including intricate curly hair. For more results please refer to the supplementary materials. }
        \vspace{-3mm}
		\label{fig:demo}
		\vspace{-2mm}
\end{figure*}

\begin{figure}[b]
		\centering
		\includegraphics[width=\linewidth]{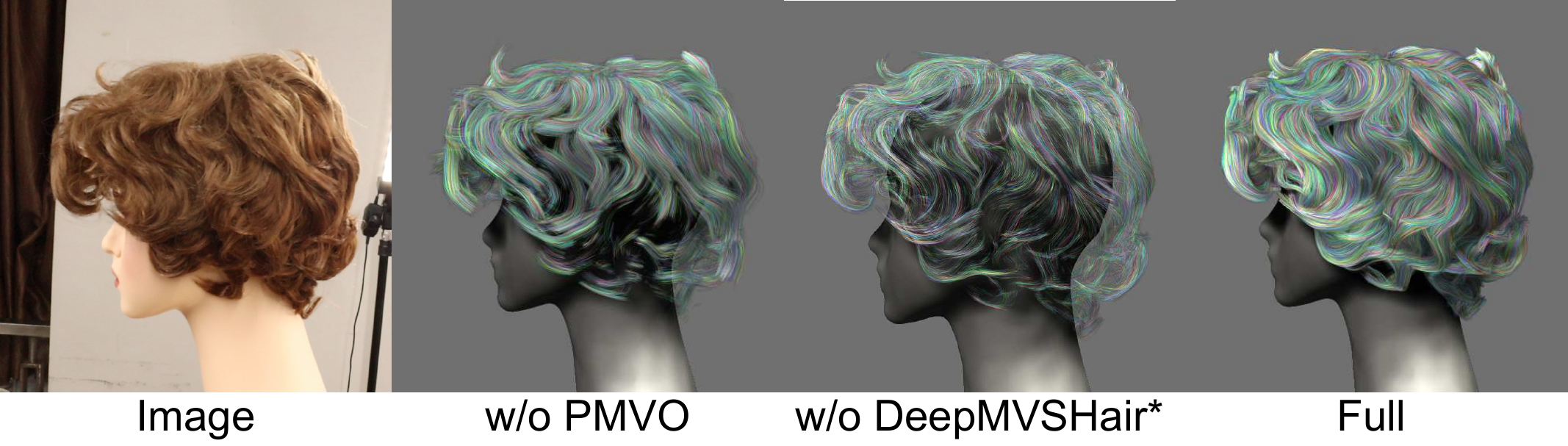}
		\caption{Qualitative evaluation of each key component of our method. \methodname~helps produce a high-quality exterior hair geometry. While DeepMVSHair* helps infer the internal geometry %
  to obtain a complete hair geometry.}
		\vspace{-3mm}
		\label{fig:ablation}
		\vspace{-4mm}
\end{figure}

Qualitative comparison with \textbf{Neural Haircut} \cite{sklyarova2023neural_haircut} is shown in \cref{fig:Compare}. Their approach tends to yield straight strands and lacks the ability to effectively represent curly hair, primarily due to the imposition of overly strong constraints by the data prior. Besides, the hair geometry representation using SDF also limits some curly strands into a smooth surface (see below for more discussion). In contrast, it is evident that our method exhibits greater robustness for curly hair. We also conduct a quantitative comparison with them \cite{sklyarova2023neural_haircut} on the %
synthetic dataset \cite{yuksel2009hair}. The comparison results are shown in \cref{tab:comparison}. Our method achieves the highest precision and F-score. %
Moreover,
we also compare the two methods in terms of reconstruction efficiency as shown in Tab.\ref{tab:time_compare}. \textbf{Neural Haircut} takes 3-4 days for each subject on a single NVIDIA RTX 3090, which significantly limits its practical application, while ours only takes 4-6 hours.

We provide %
qualitative comparisons with \textbf{Instant-NGP}, \textbf{NeuS}, and \textbf{NeuS-\methodname}. \textbf{Instant-NGP} and \textbf{NeuS} %
can only produce a coarse hair geometry, our method can achieve more robust and accurate results than them. It is important to note that the results obtained by NeRF-based methods often exhibit lots of noise, and \textbf{NeuS} is more effective in obtaining clean 3D geometry. However, as shown in the last two columns of %
\cref{fig:compare_neus}, the results of \textbf{NeuS-\methodname} are smoother in some details, and thus we choose \textbf{Instant-NGP} %
to initialize the coarse geometry.

Quantitative and qualitative comparisons with \textbf{DeepMVSHair} are given in \cref{tab:comparison} and \cref{fig:compare_mvs}, respectively. While their method is capable of generating plausible geometry, its performance is significantly impacted by the quality of the input 2D orientation map. On the other hand, taking in undirectional strand maps derived from the 3D line map, the proposed \textbf{DeepMVSHair*} %
can produce a better result. However, since the learning method is limited by the distribution of the data prior,  the obtained results may differ from the captured images in detail. In contrast, our method synthesizes a high-quality {exterior} structure of the hair and only applies data priors to the invisible inner geometry, resulting in superior results.

As shown in \cref{fig:demo}, we also provide some challenging cases of real-world video capture to evaluate our method. Our method can robustly reconstruct diverse hairstyles, including straight, wavy, and curly hair, and achieve high-quality and realistic %
results. For more examples, please refer to our supplementary materials.

\subsection{Ablation Study}\label{ablation}
We evaluate the performance of each component of our method via %
an ablation study on real data and synthetic data \cite{yuksel2009hair}. As shown in \cref{fig:ablation}, without PMVO, errors in camera parameters lead to the removal of many hair points as noise, resulting in the loss of some hair strands. On the other hand, when DeepMVSHair* is not applied, the results lack internal structures, appearing as a shell comprised of isolated segments. For more ablation studies please refer to our supplementary materials.



\section{Conclusion and Discussion} %
\label{limitation}
In this paper, we have rethought %
the existing multi-view based hair reconstruction pipeline, where most methods apply the learned data prior directly to the reconstruction of the entire hairstyle, which is extremely limited by the diversity of training databases. To this end, {we proposed \sysname, a generic framework that bifurcates hair modeling into exterior and interior geometries. It extracts the exterior hair structure from multiview images without relying on data priors. Subsequently, the framework deduces the inner hair structure by combining learned data priors with the extracted high-quality exterior hair structure.}
Extensive experiments demonstrated that our method employing coarse geometry produced by \cite{muller2022instant} combined with the proposed \methodname~and inner inference module can %
can reconstruct a high-fidelity strand model and support various hairstyles, including curly hair.

\begin{figure}[ht]
		\centering
		\includegraphics[width=\linewidth]{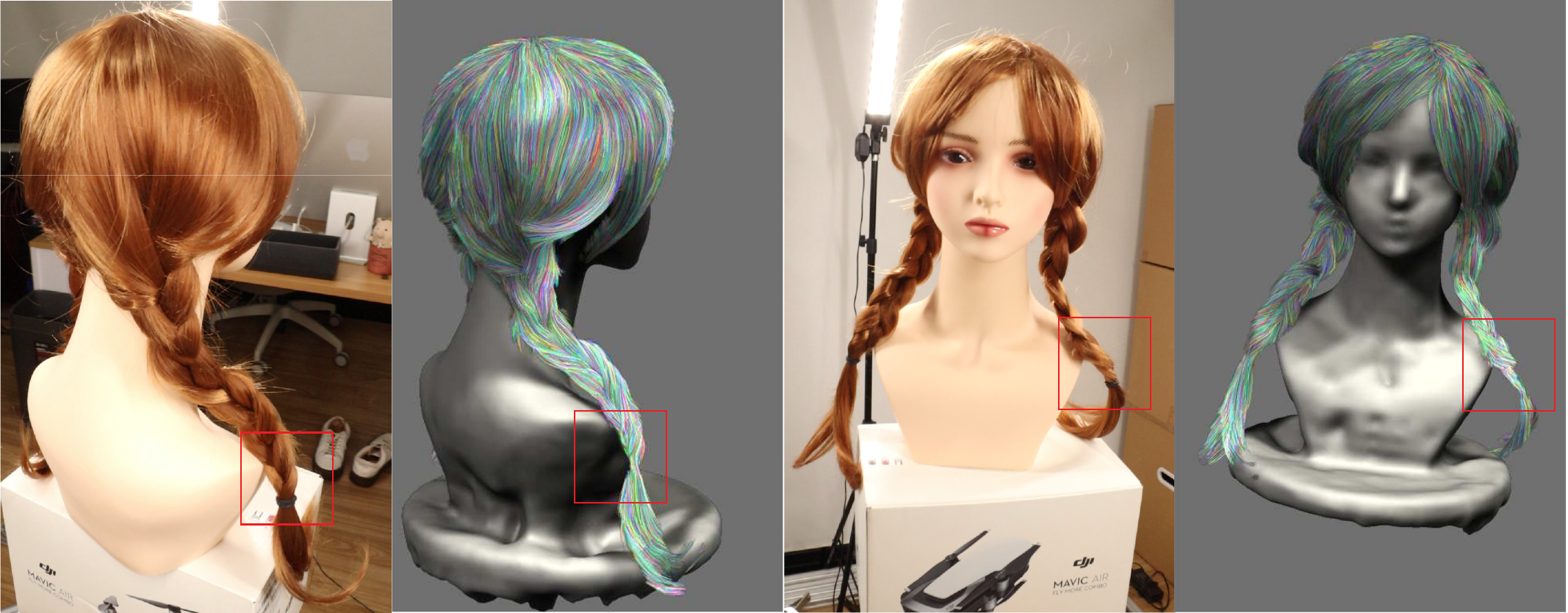}
		\caption{The main limitation of our method is that the connection relationships may be incorrect in instances with severe intersections and occlusions.}
		\label{fig:limitation}
		\vspace{-5mm}
\end{figure}

As shown in \cref{fig:demo} and \cref{fig:limitation}, the main limitation of our method is that, while we can successfully reconstruct the majority of hair geometry, due to severe intersections and occlusions, some intricate hairstyles (such as braids), the connection relationships may be incorrect. This is primarily because despite we can reconstruct high-quality hair exterior, %
the inner geometry remains dependent on the data prior. In principle, expanding the dataset to include a more diverse range of hairstyles {or directly reconstructing the interior structure using computed tomography similar to CT2Hair \cite{shen2023ct2hair}} 
can alleviate this limitation.

\section*{Acknowledgements}
This work is supported in part by the NSF China (No. 62172363). Besides, We thank the State Key Lab of CAD\&CG for providing the computational resources for this work. We also sincerely thank Vanessa Sklyarova for providing the data and reconstruction results to compare with Neural Haircut.

{
    \small
    \bibliographystyle{ieeenat_fullname}
    \bibliography{main}
}
\maketitlesupplementary

\section{Implementation Details and Data Preprocessing}
\subsection{Datasets preprocessing}

\paragraph{Synthetic data.}  We train the DeepMVSHair* on USC-HairSalon\cite{hu2015single}, which contains 343 hairstyles aligned with a template head. We augment the dataset to 2,744 samples with three degrees of freedom: scaling, rotation, and translation. Then, for each sample, we generate a strand model, the ground-truth 3D orientation, and a rendered 2D unidirectional map with 16 fixed views. We also evaluate our method on another synthetic dataset \cite{yuksel2009hair}. Specifically, {for each model}, we use Blender \cite{Blender} to render 150 RGB images with a resolution of $1920\times1080$, employing different camera poses. Then, we reconstruct the corresponding strand model 
and evaluate it with the ground truth.

\paragraph{Real-world data.} We also evaluate our method utilizing a public multi-view H3DS dataset \cite{yuksel2009hair} and {a set of} real-world captured monocular videos. For {the} H3DS dataset, each scene has 32 views {with their} %
corresponding camera parameters. For monocular videos, to better initialize the coarse geometry, we select 150-200 frames around the subject using an image quality assessment network \cite{su2020blindly} and employ structure-from-motion with COLMAP \cite{schonberger2016structure} to obtain the camera intrinsic and extrinsic parameters. Then, we calculate the 2D orientation map utilizing 180 Gabor filters with $\sigma_{x}=1.8$, $\sigma_{y}=2.4$, frequency $\omega=0.25$, and kernel size $k=17$. Lastly, we also generate hair masks using human matting \cite{ke2022modnet} and semantic segmentation \cite{liu2022cdgnet}.

\subsection{Implementation details}
To initialize the coarse point cloud, we densely sample around the coarse geometry produced by Instant-NGP \cite{muller2022instant} at a high resolution of $512\times512\times384$. For the PMVO, we remove the low confidence area when performing {the} patch sample strategy. Besides, we select 10 reference frames based on confidence values for different initializations of $\LineP$ to calculate $\mathcal{L}_{opt}$ more robustly. For interior hair geometry inference, we grow short segments using $\LineMap$ to render the undirectional strand maps. Then, we train DeepMVSHair* on 16 fixed view{s} at $1280\times720$ resolution with the learning rate {of} 0.0005. The training process takes about 6 days on a single NVIDIA RTX 3090, 3 days for {the} 3D orientation prediction network and 3 days for the 3D occupancy prediction network.

\section{Additional Experiments and Results}

\subsection{More comparisons}

\begin{figure}[t]
		\centering
		\includegraphics[width=\linewidth]{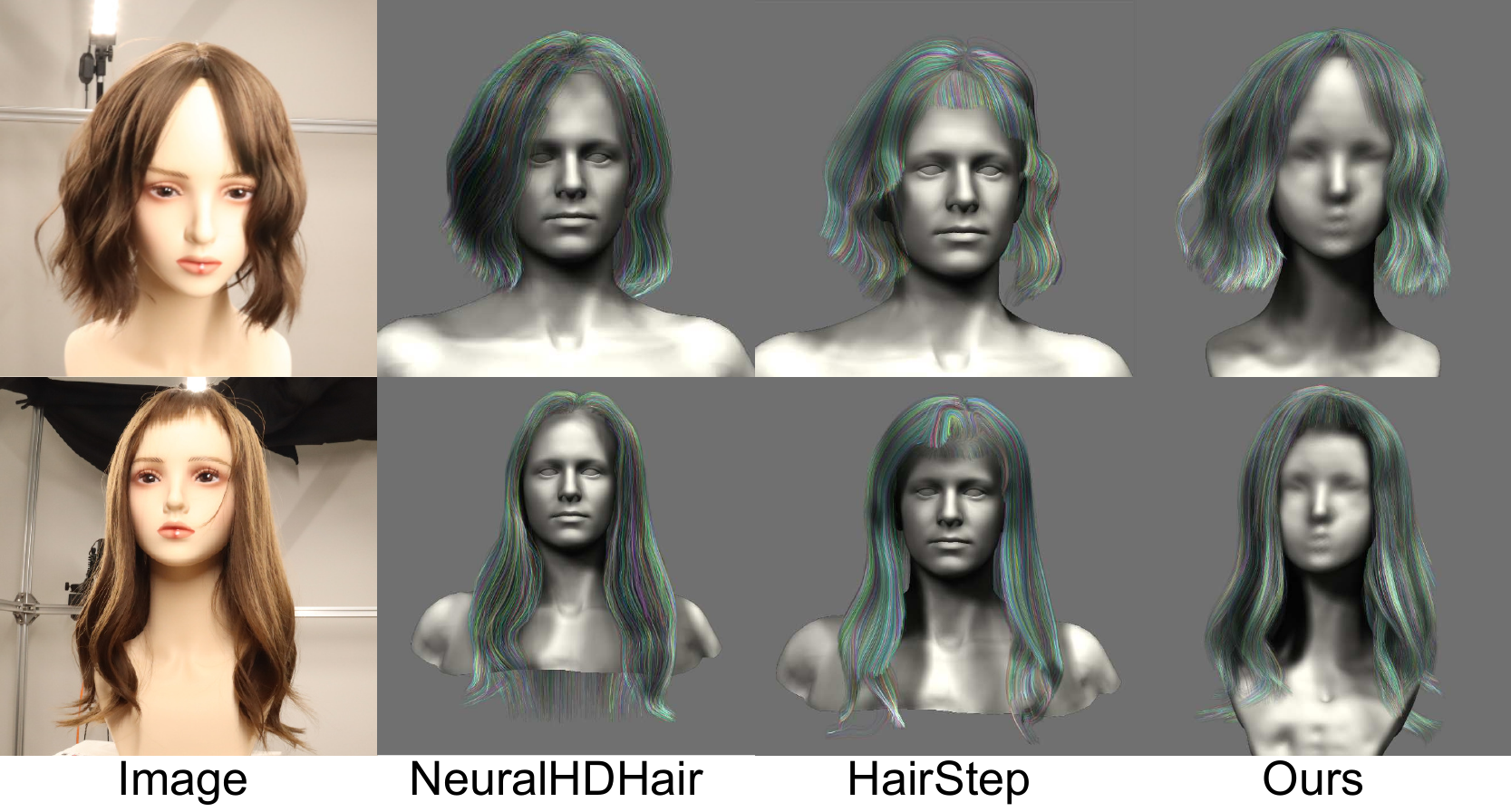}
		\caption{Qualitative comparison with single view-based hair modeling methods. The reconstruction {results are} produced by the open source code and pre-trained models.}
        \vspace{-2mm}
		\label{fig:compare_single}
        \vspace{-3mm}

\end{figure}

\paragraph{Comparison with single {view-based} image method{s}.} We provide {an} extended qualitative comparison with the single view-based reconstruction methods \cite{wu2022neuralhdhair,zheng2023hairstep}. As shown in \cref{fig:compare_single}, obviously, our method {achieves} %
higher fidelity results. 

\begin{figure*}[t]
		\centering
		\includegraphics[width=0.8\textwidth]{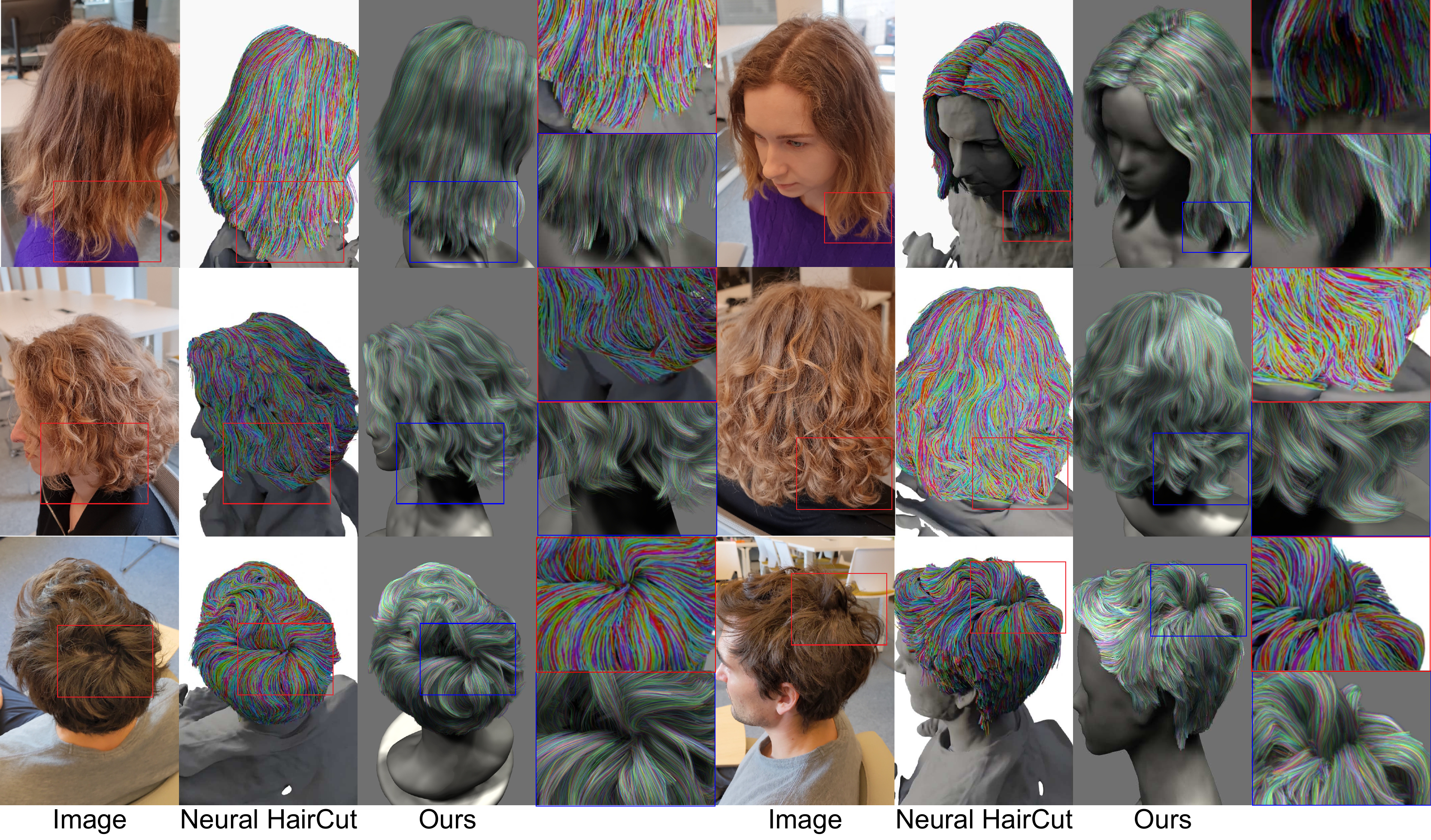}
		\caption{Extended qualitative comparison with Neural Haircut \cite{sklyarova2023neural_haircut}.}
		\label{fig:supple_compare}
        \vspace{-3mm}

\end{figure*}

\paragraph{Extended comparison with Neural Haircut.} %
We provide {an} additional comparison with Neural HairCut \cite{sklyarova2023neural_haircut} {in \cref{fig:supple_compare}}. While their method can handle both long and short hair, it struggles to {accurately} reconstruct some complex hairstyles, such as curly hair.  %
In contrast, our approach is versatile and adaptable to more hairstyles.

\subsection{Additional ablation studies}

\begin{figure}[t]
		\centering
		\includegraphics[width=\linewidth]{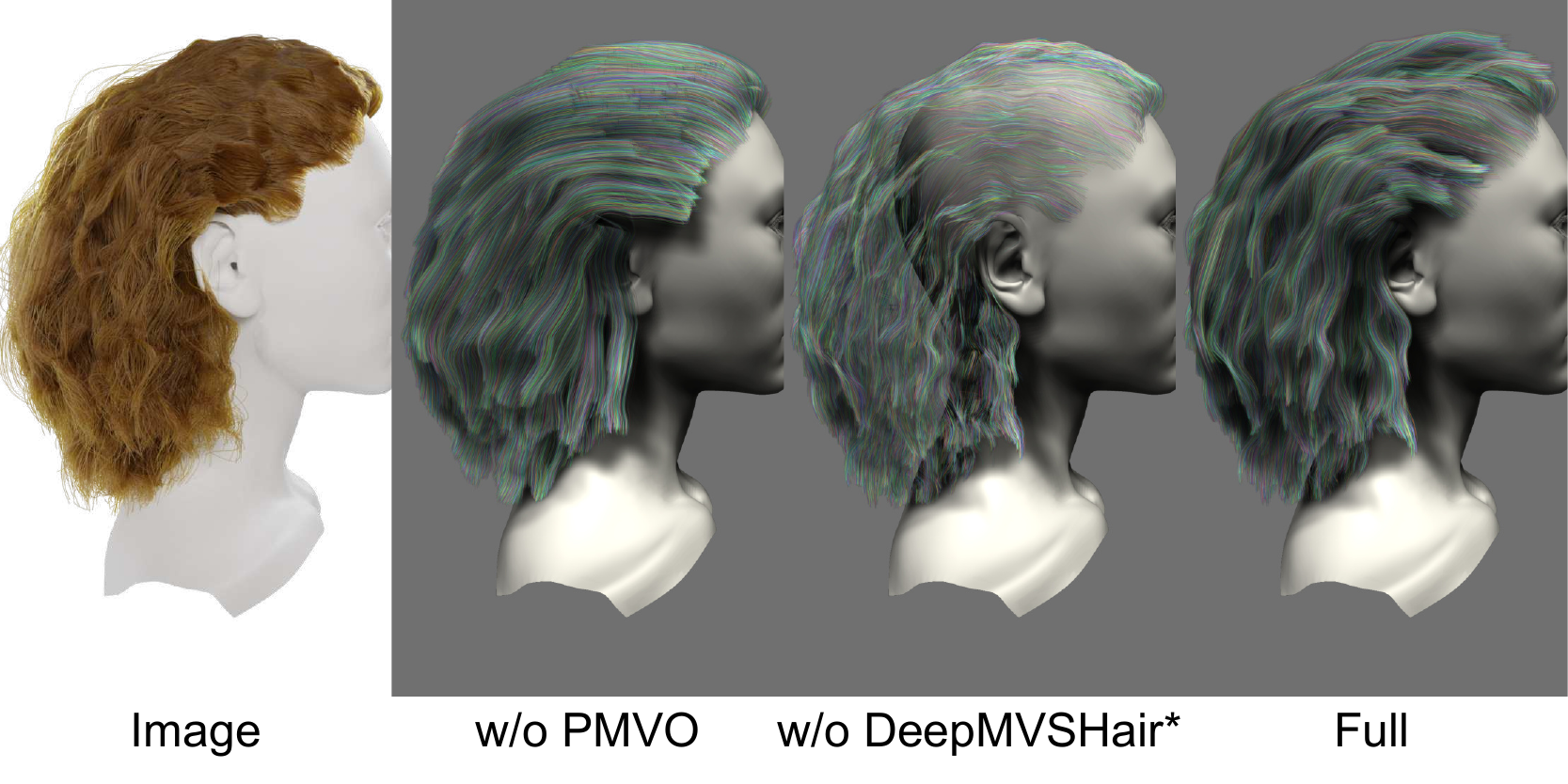}
		\caption{Qualitative evaluation of each key component of our method on synthetic data.}
		\label{fig:qualitative_ablation}
        \vspace{-2mm}
\end{figure}

\paragraph{Ablation studies on synthetic data.} We also evaluate each component of our method on {a} synthetic dataset \cite{yuksel2009hair}. We render the ground-truth 3D strand model 
into RGB images using Blender and reconstruct it. As shown in \cref{fig:qualitative_ablation}. PMVO helps us maintain high-fidelity exterior hair geometry while DeepMVSHair* assists in reconstructing the complete strand model. As indicated in Tab. \ref{tab:ablation_metric}, without \methodname, both precision and recall are greatly reduced due to the errors of camera parameters. On the other hand, without DeepMVSHair*, we can achieve the highest precision but the lowest recall, since the reconstruction results contain only the outer layer of hair. Finally, our full method combining \methodname~and DeepMVSHair* achieves %
the highest recall and F-score.

\begin{figure}[t]
		\centering
		\includegraphics[width=\linewidth]{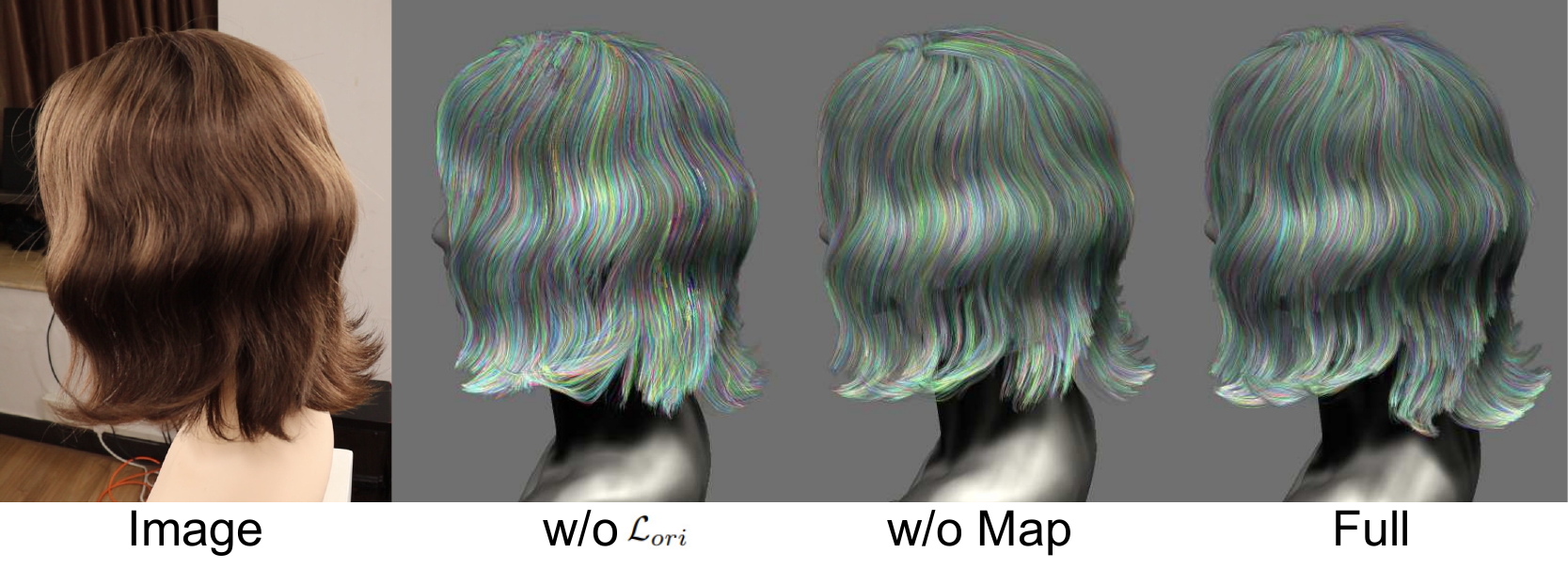}
		\caption{Qualitative evaluation of the improvements to DeepMVSHair.}
		\label{fig:evaluate_deepmvshair}
        \vspace{-2mm}
\end{figure}

\paragraph{Evaluation of DeepMVSHair*.} To evaluate the effectiveness of our improvements to DeepMVSHair, we conducted two sets of ablation studies: 1) without the undirectional strand map (w/o Map) and 2) without the proposed new loss function $\mathcal{L}_{ori}$. The comparison results are shown in \cref{fig:evaluate_deepmvshair}. With no undirectional strand map, the reconstructed results are more susceptible to the loss of geometric details due to ambiguity. While the loss function $\mathcal{L}_{ori}$ helps us train DeepMVSHair* more robustly.

\begin{figure*}[h]
		\centering
		\includegraphics[width=\textwidth]{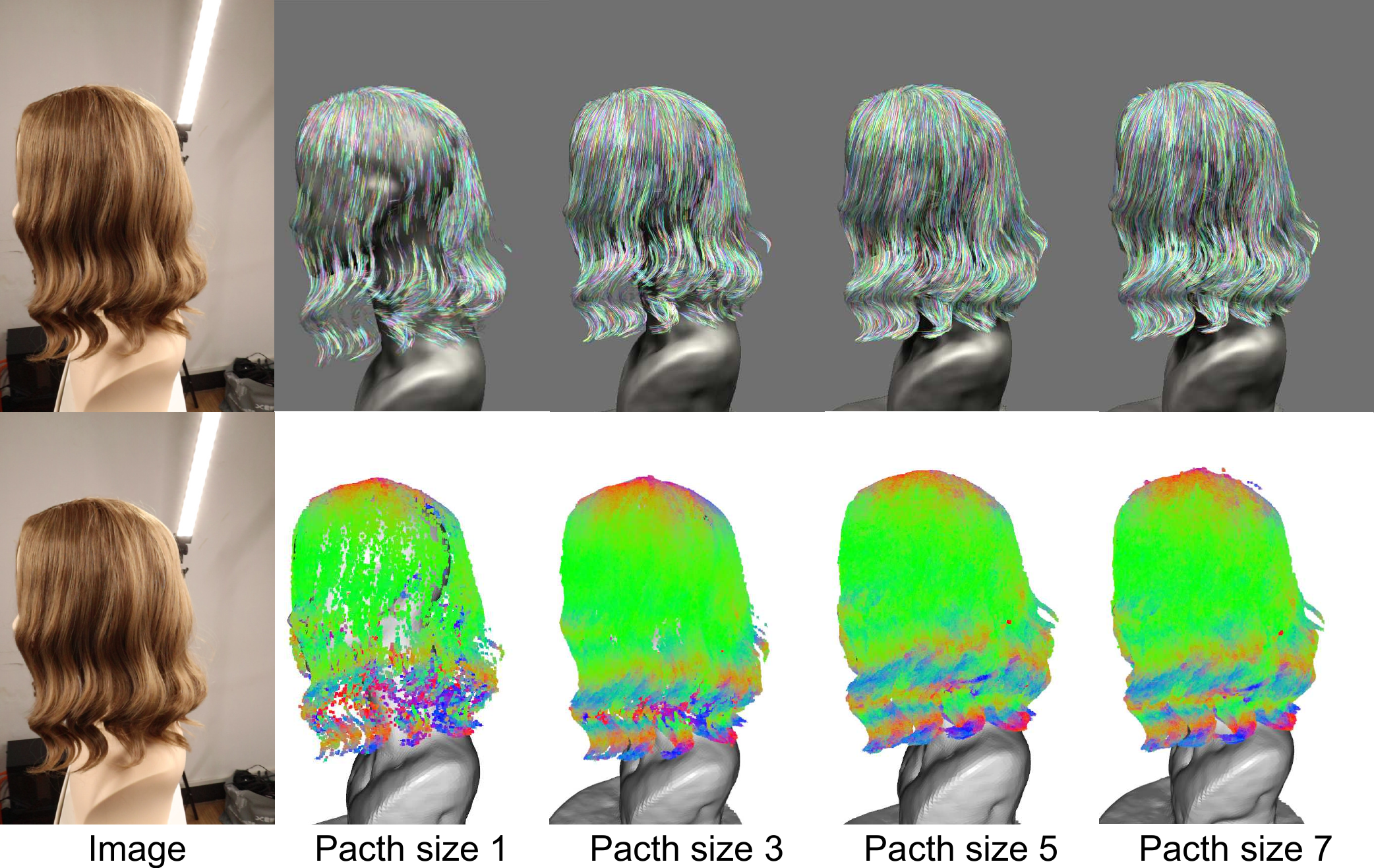}
		\caption{Qualitative comparison of \methodname~with different patch sizes. When the patch size is greater than 5, there is no significant improvement in the reconstructed geometry. The above results are all the exterior geometry of the hair.}
		\label{fig:patch_size}

\end{figure*}

\paragraph{Comparison with different patch size{s}.} We also compared the effects of different sample sizes (patch size) in \methodname~ on the reconstruction of the exterior hair geometry. As shown in \cref{fig:patch_size} and \cref{tab:patch_size}{, setting} %
the patch size to 5 %
better balance{s} performance and efficiency.

\begin{figure}[t]
    \centering
    \resizebox{\linewidth}{!}{
    \begin{tabular}{l rrr | rrr | rrr}
        \setlength{\tabcolsep}{0pt}
        & \multicolumn{9}{c}{\textbf{Thresholds: mm} $/$ \textbf{degrees}} \\
        \textbf{Method} & $2 / 20$ & $3 / 30$ & $4 / 40$ & $2 / 20$ & $3 / 30$ & $4 / 40$ & $2 / 20$ & $3 / 30$ & $4 / 40$ \\
        \cline{2-10}
        & \multicolumn{3}{c}{\textbf{Precision}} & \multicolumn{3}{c}{\textbf{Recall}} & \multicolumn{3}{c}{\textbf{F-score}} \\
        \hline
        w/o PMVO	&	30.3	&	54.5	&	67.3	&	6.7	&	14.5	&	21.6	&	11.0	&	22.9	&	32.7	\\
 w/o DeepMVSHair*	&	\textbf{65.2}	&	\textbf{87.4}	&	\textbf{95.6}	&	6.3	&	15.2	&	19.8	&	11.5	&	25.9	&	32.8	\\
 Full	&	60.8	&	83.3	&	92.1	&	\textbf{10.4}	&	\textbf{19.3}	&	\textbf{25.9}	&	\textbf{17.8}	&	\textbf{31.3}	&	\textbf{40.4}	\\

    \end{tabular}
    }
    \vspace{-0.1cm}
    \setlength{\belowcaptionskip}{-3mm}
    \captionof{table}{Quantitative evaluation of individual components of our method on the %
    synthetic dataset \cite{yuksel2009hair}.}
    \label{tab:ablation_metric}
\end{figure}

\subsection{More results}
\begin{figure*}[t]
		\centering
		\includegraphics[width=\textwidth]{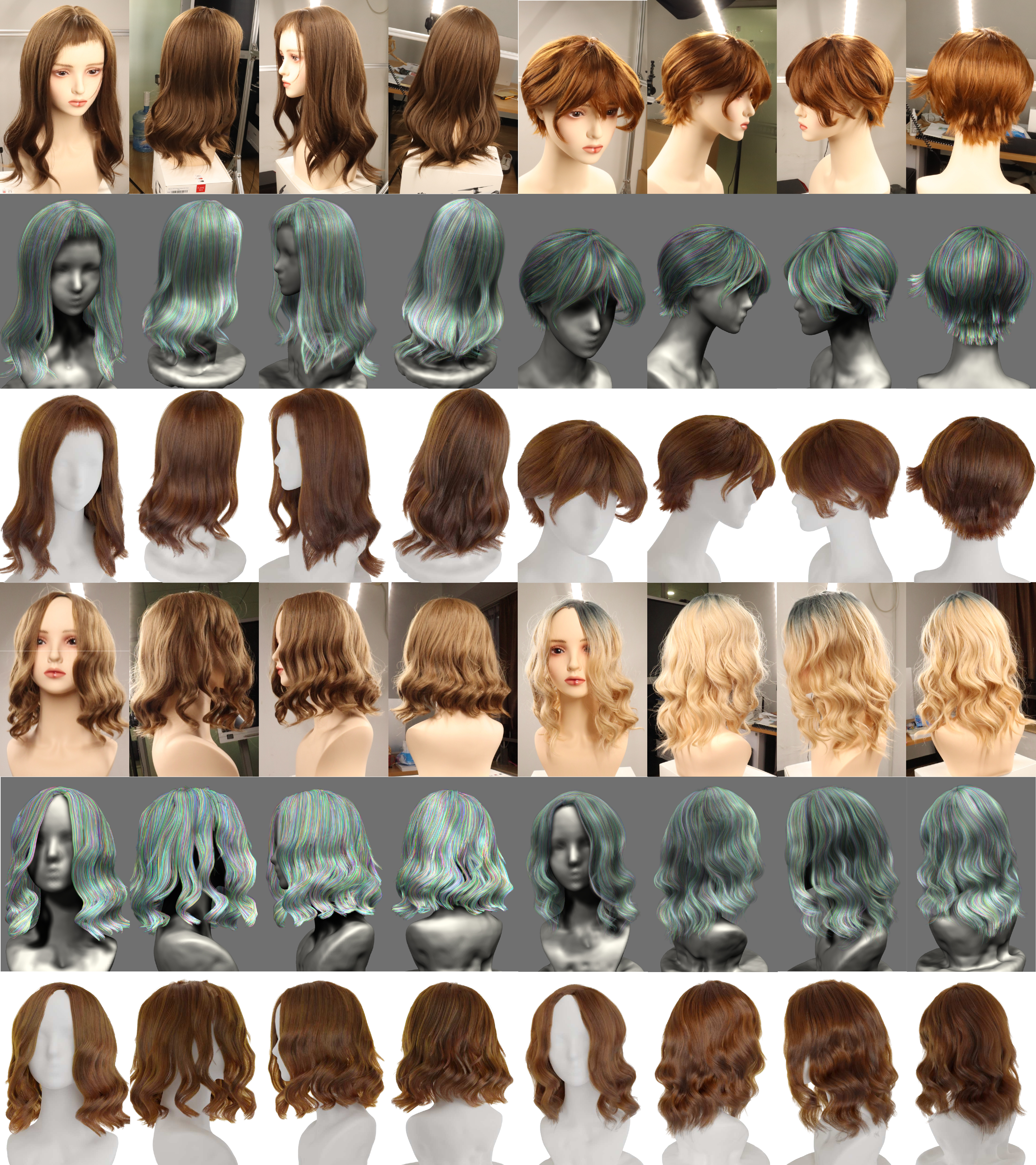}
		\caption{Additional reconstruction results of our method based on monocular videos. We use two rendering styles to showcase our results: colorful strands for displaying hair geometry and Blender \cite{Blender} for realistic rendering.}
		\label{fig:more_demo3}

\end{figure*}

\begin{figure*}[t]
		\centering
		\includegraphics[width=0.85\textwidth]{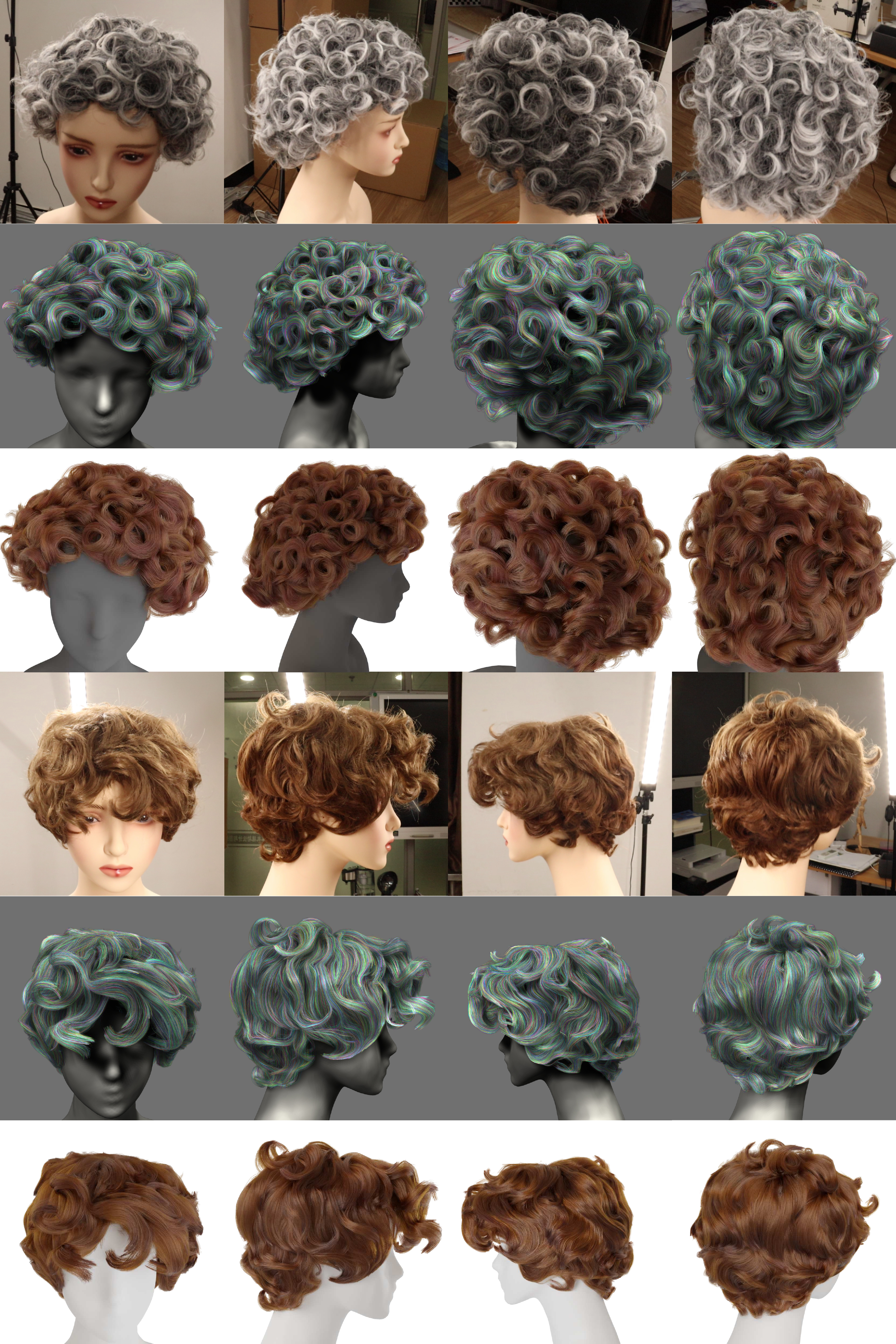}
		\caption{Additional reconstruction results of our method based on monocular videos.}
		\label{fig:more_demo1}

\end{figure*}

\begin{figure*}[t]
		\centering
		\includegraphics[width=\textwidth]{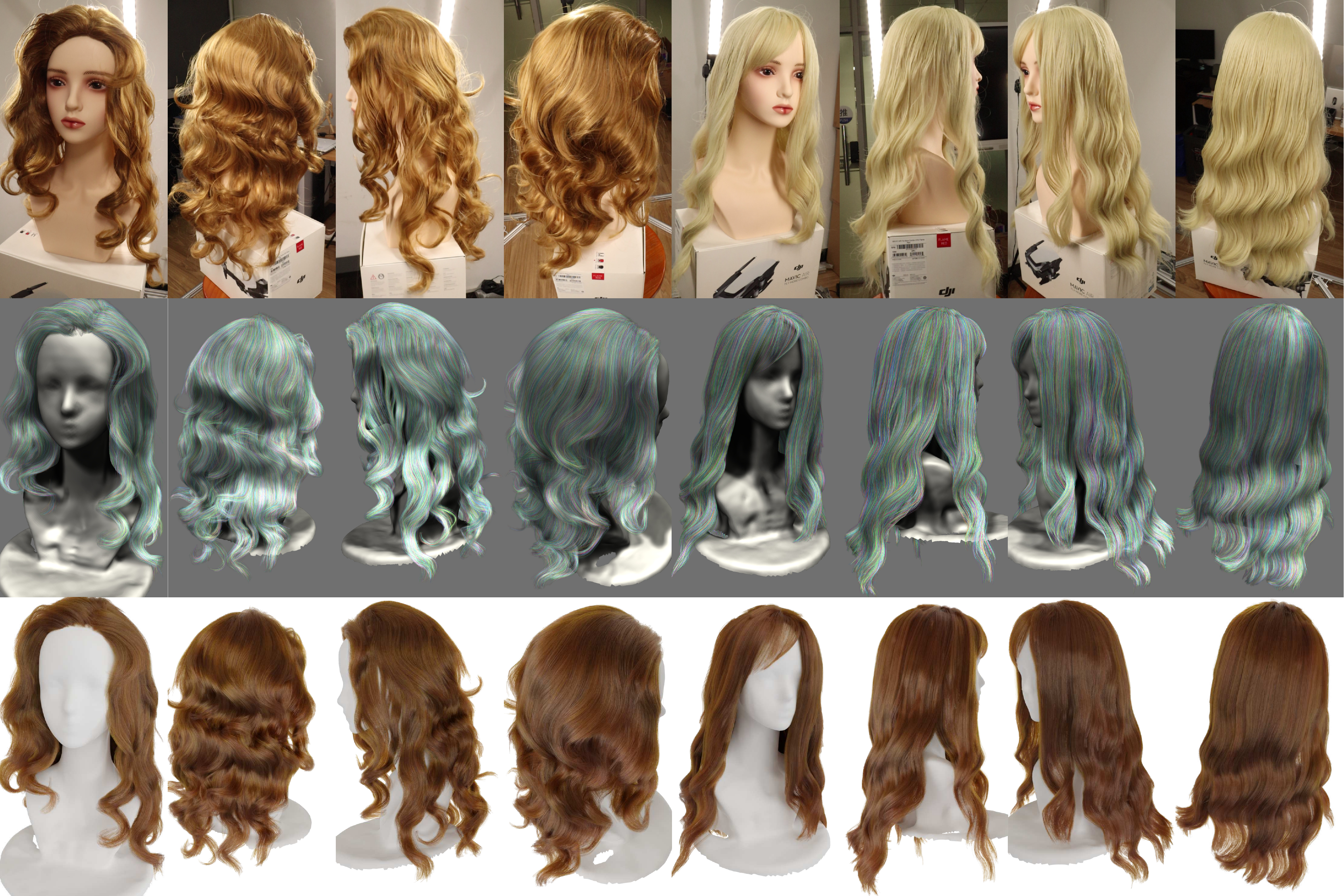}
		\caption{Additional reconstruction results of our method based on monocular videos.}
		\label{fig:more_demo2}

\end{figure*}

Lastly, we provide additional reconstruction examples from a monocular video showcasing various hairstyles such as short, long, straight, curly, and wavy hair. As shown in \cref{fig:more_demo3,fig:more_demo1,fig:more_demo2}, \sysname~can handle diverse, complex hairstyles and achieve high-fidelity results.

\begin{figure}
    \centering
        \resizebox{\linewidth}{!}{
        \begin{tabular}{c|c| c| c| c|}
        \hline
            Patch size & 1 & 3 & 5 & 7  \\
          \hline
            Time & $\sim$30min & $\sim$ 90min & 3h-4h & 6h-8h \\  \hline
            
        \end{tabular}
        }      
 
    \captionof{table}{Comparison of time consumption under different patch sizes. As the patch size increases, the time consumption  increases significantly.}
    \label{tab:patch_size}
    
\end{figure}

\end{document}